\documentclass{article}
\usepackage{placeins}
\usepackage{microtype}
\usepackage{graphicx}
\usepackage{subfigure}
\usepackage{booktabs} % for professional tables

\usepackage{hyperref}

\usepackage[accepted]{arxiv}

\usepackage{amsmath}
\usepackage{amssymb}
\usepackage{mathtools}
\usepackage{amsthm}

\usepackage[capitalize,noabbrev]{cleveref}

\theoremstyle{plain}

\theoremstyle{definition}

\theoremstyle{remark}

\usepackage[textsize=tiny]{todonotes}

\usepackage{hyperref}
\usepackage{url}
\usepackage{tabularx} % For adjustable width tables
\usepackage{makecell}
\usepackage{tikz}
\usetikzlibrary{shadows}

\usepackage{wrapfig}
\usepackage{listings}
\usepackage{enumitem}

\usepackage{bm}
\usepackage{bbm}
\usepackage{authblk}

\usepackage{multirow}

\icmltitlerunning{Distill Not Only Data but Also Rewards:  Can Smaller Language Models Surpass Larger Ones?}
\begin{document}

\twocolumn[
\icmltitle{Distill Not Only Data but Also Rewards: \\ Can Smaller Language Models Surpass Larger Ones?}



\icmlsetsymbol{equal}{*}

% \begin{icmlauthorlist}
% \icmlauthor{Yudi Zhang}{tue}
% \icmlauthor{Lu Wang}{msra}
% \icmlauthor{Meng Fang}{liverp}
% \icmlauthor{Yali Du}{kcl}
% \icmlauthor{Chenghua Huang}{fudan}
% \icmlauthor{Jun Wang}{ucl}
% \icmlauthor{Qingwei Lin}{msra}
% \icmlauthor{Mykola Pechenizkiy}{tue}
% \icmlauthor{Dongmei Zhang}{msra}
% \icmlauthor{Saravan Rajmohan}{msra}
% \icmlauthor{Qi Zhang}{msra}

% \icmlaffiliation{tue}{Eindhoven University of Technology}
% \icmlaffiliation{msra}{Microsoft Asia}
% \icmlaffiliation{liverp}{University of Liverpool}
% \icmlaffiliation{kcl}{King’s College London}
% \icmlaffiliation{fudan}{Fudan University}
% \icmlaffiliation{ucl}{University College London}
% %\icmlaffiliation{intern}{Work done during Yudi's internship in Microsoft.}
% \end{icmlauthorlist}

\begin{center}
\textbf{
Yudi Zhang\textsuperscript{1}, 
Lu Wang\textsuperscript{2}, 
Meng Fang\textsuperscript{3,1}, 
Yali Du\textsuperscript{4}, 
Chenghua Huang\textsuperscript{5}, 
Jun Wang\textsuperscript{6}, 
Qingwei Lin\textsuperscript{2}, 
Mykola Pechenizkiy\textsuperscript{1}, 
Dongmei Zhang\textsuperscript{2}, 
Saravan Rajmohan\textsuperscript{2}, 
Qi Zhang\textsuperscript{2}
}

\vspace{0.5em} % 控制作者与单位的间距

\textsuperscript{1} Eindhoven University of Technology, 
\textsuperscript{2} Microsoft, 
\textsuperscript{3} University of Liverpool, 
\textsuperscript{4} King’s College London, 
\text{\textsuperscript{5} Fudan University}, 
\textsuperscript{6} University College London
\end{center}

% \affil[1]{Eindhoven University of Technology}
% \affil[2]{Microsoft}
% \affil[3]{Fudan University}
% \affil[4]{University of Liverpool}
% \affil[5]{King’s College London}

% \icmlcorrespondingauthor
\icmlcorrespondingauthor{Lu Wang}{wlu@microsoft.com}
\icmlcorrespondingauthor{Meng Fang}{Meng.Fang@liverpool.ac.uk}

\icmlkeywords{Machine Learning, ICML}

\vskip 0.3in]


\printAffiliationsAndNotice  % leave blank if no need to mention equal contribution

\begin{abstract}

Distilling large language models (LLMs) typically involves transferring the teacher model's responses through supervised fine-tuning (SFT). However, this approach neglects the potential to distill both data (output content) and reward signals (quality evaluations). Extracting reliable reward signals directly from teacher models is challenging, as LLMs are optimized for generation rather than evaluation, often resulting in biased or inconsistent assessments. To address this limitation, we propose a novel distillation pipeline that transfers both responses and rewards. Our method generates pseudo-rewards through a self-supervised mechanism that leverages the inherent structure of both teacher and student responses, enabling reward learning without explicit external evaluation.
The reward model subsequently guides reinforcement learning (RL), allowing iterative refinement of the student model after an SFT warm-up phase.
Experiments on GSM8K and MMLU-PRO demonstrate that our method consistently outperforms traditional SFT-based approaches, enabling student models to surpass the performance of their teachers. This work highlights the potential for scalable, efficient distillation through structured self-supervised reward learning, reducing dependence on external reward supervision.

\end{abstract}

\vspace{-2pt}
\section{Introduction}

\begin{figure}[!ht]
    \centering
    \includegraphics[width=0.9\linewidth]{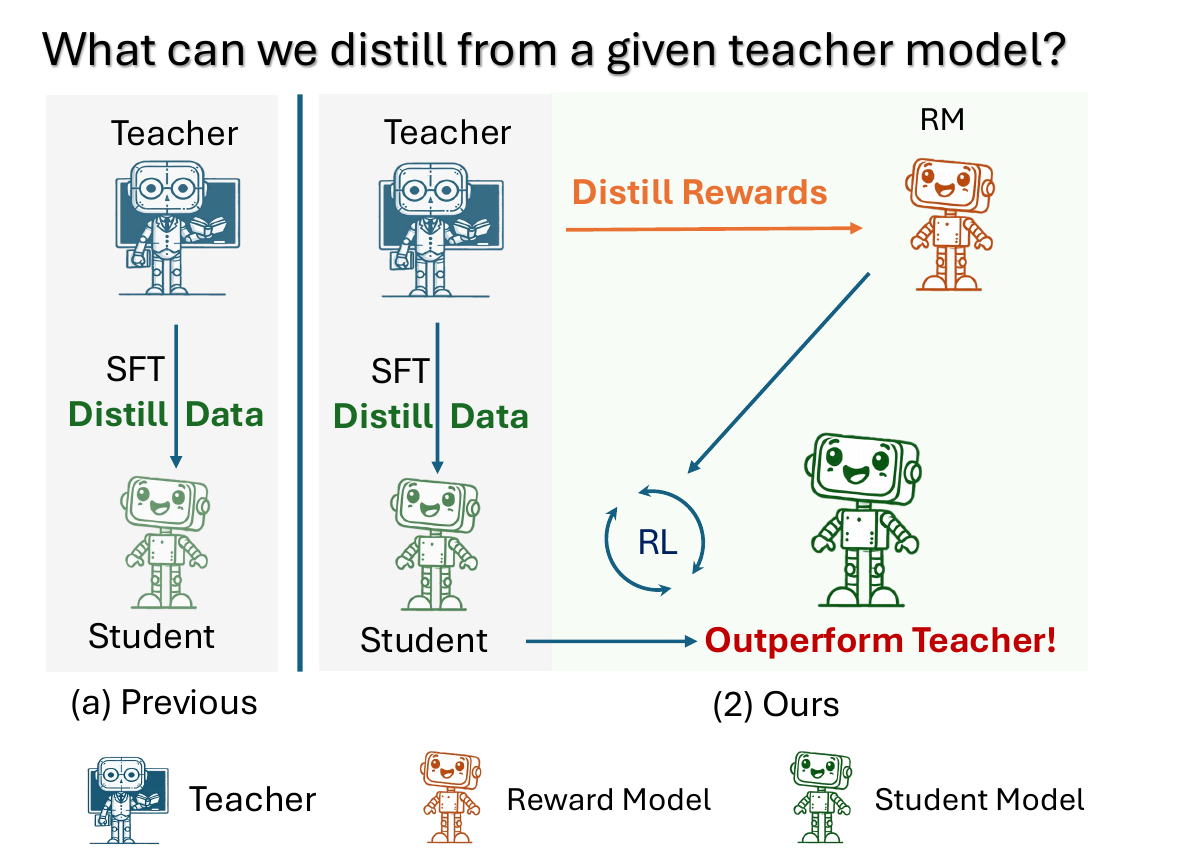}
    \vspace{-10pt}
    \caption{Comparison of (a) traditional knowledge distillation approaches using supervised fine-tuning (SFT) to distill data directly into a student model, and (b) our proposed method, which distills both data (output content) and rewards (quality evaluations) in a self-supervised manner. By training a reward model (RM) and applying reinforcement learning (RL), the student model progressively surpasses the teacher's performance.}
    \label{fig:fig1}
    \vspace{-10pt}
\end{figure}

\begin{figure*}[!ht]
    \centering
    \includegraphics[width=0.85\linewidth]{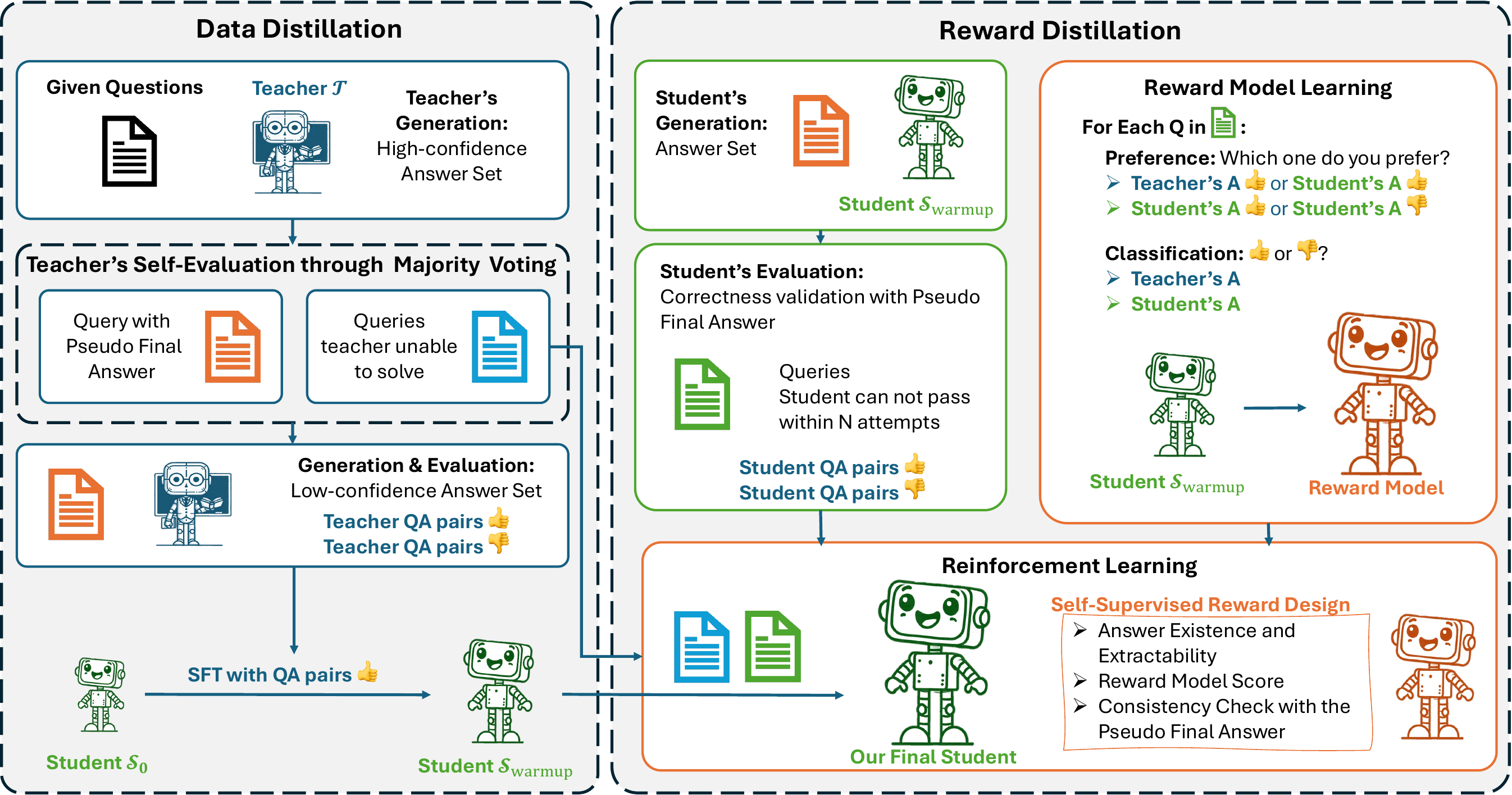}
    \caption{The overall framework of the proposed method. \textbf{Data Distillation:} Teacher LLMs first generate and self-evaluate their answers, then select high-quality responses for the student model warm-up via supervised fine-tuning (SFT). \textbf{Reward Distillation:} The student is further refined through reinforcement learning using a self-supervised reward mechanism.}
    \label{fig:framework}
    \vspace{-5pt}
\end{figure*}

% \FloatBarrier
\begin{table*}[!ht]
\small
\centering
\caption{Example of a question and the teacher's response. The final answer is shown in \textcolor{blue}{blue}.}
\begin{tabularx}{\linewidth}{@{}l X@{}}
\toprule
\textbf{Q} & 
Weng earns \$12 an hour for babysitting. Yesterday, she just did 50 minutes of babysitting. How much did she earn? \\ \midrule
\textbf{A} & 
Let's break this down step by step! Step 1:
50 minutes is equal to $\frac{50}{60} = \frac{5}{6}$ of an hour. Step 2: Multiply her hourly rate by the fraction of an hour she worked: $\$12 \times \frac{5}{6} = \$10$. The answer is \textcolor{blue}{\$10}.
\\ \bottomrule
\end{tabularx}
\label{tab:qa_example}
\vspace{-10pt}
\end{table*}
Knowledge Distillation has emerged as a promising technique for mitigating the high computational demands of Large Language Models (LLMs) by training smaller student models under the guidance of larger teacher models. Achieving competitive performance through fine-tuning, however, critically depends on high-quality annotated data—a resource that remains a significant bottleneck~\citep{achiam2023gpt, kaplan2020scaling, roziere2023code, yuan2023scaling, luo2023wizardmath}. Supervised fine-tuning (SFT) using distilled data has become a prevalent approach~\cite{feng2021survey}, where teacher LLMs generate responses for various tasks or domains, sometimes in conjunction with search and selection strategies~\cite{tian2024toward, zhang2024rest}.

Although SFT-based methods~\citep{magister2023teaching, fu2023specializing} provide a straightforward distillation process, 
% they fail to harness the full potential of teacher LLMs, particularly their capacity to implicitly signal the quality of generated responses. \mf{=
they do not fully exploit teacher LLMs' inherent potential, particularly their ability to implicitly convey the quality of generated responses.
% } 
Techniques like Reinforcement Learning from Human Feedback~\cite{wang2024comprehensive} and reward-based training~\cite{setlur2024rewarding, wang2024math} have demonstrated that reward models, which quantify response quality, can significantly enhance model performance. However, these methods typically require substantial human effort to produce reliable reward signals. 
While teacher LLMs can provide both responses and evaluations, directly prompting them to evaluate their own outputs is unreliable. This unreliability stems from their primary optimization for text generation, leading to biased or inconsistent evaluations due to issues like hallucination~\cite{huang2023survey} and inadequate uncertainty handling~\cite{xiong2023can}.
% While teacher LLMs could, in principle, provide both responses and evaluations, directly prompting them to evaluate their own outputs is unreliable: LLMs are primarily optimized for text generation, often producing biased or inconsistent evaluations due to limitations such as hallucination~\cite{huang2023survey} and inadequate uncertainty handling~\cite{xiong2023can}.

In this paper, we propose a novel distillation framework that eliminates the need for any external reward signals from the teacher. Instead, our method generates ``pseudo-rewards'' by exploiting inherent structures and relationships within the teacher's response data. This approach allows the model to infer reliable reward signals without relying on direct or explicit teacher evaluations, which are often biased and inconsistent.

To address the challenges of distribution shift and reward alignment, our framework considers both the teacher's and the student's responses when generating these pseudo-rewards. 
% By comparing and contrasting the quality of responses from both models, the reward model learns to prioritize high-quality outputs while adapting dynamically to the evolving performance of the student. 
By comparing and contrasting the quality of responses from both models, the reward model learns to prioritize high-quality outputs while adapting to the student‘ performance. This reward-driven refinement process improves the student model through reinforcement learning (RL) after an initial supervised fine-tuning (SFT) phase, enabling it to surpass the teacher in performance under certain conditions.
This reward-driven refinement process improves the student model through reinforcement learning (RL) after an initial supervised fine-tuning (SFT) phase, enabling it to surpass the teacher in performance under certain conditions.

Our key contributions are as follows. First, we design a new distillation pipeline that distills not only data but also rewards to enhance student model training. Our approach eliminates the need for external teacher-generated reward signals. Second, we propose a self-supervised reward learning method that generates pseudo-rewards by leveraging the inherent structure of both teacher and student responses. This enables reward learning without the need for explicit external evaluation while effectively addressing distribution shift. Third, we combine supervised fine-tuning (SFT) with reinforcement learning (RL) driven by pseudo-rewards to progressively refine the student model. Experiments on GSM8K and MMLU-PRO demonstrate that our approach surpasses traditional SFT-based methods, enabling student models to exceed the performance of their teachers.

\section{Related Work}
\label{sec:relatedwork}
In this section, we review several relevant topics of our method, including knowledge distillation and reinforcement learning from external feedback.

\textbf{Knowledge Distillation.} Recent studies on knowledge distillation for language models have primarily focused on transferring reasoning capabilities from large language models (LLMs) to smaller models~\cite{shridhar2022distilling, magister2023teaching}. For example, \citet{shridhar2022distilling} employed semantic decompositions to distill reasoning skills. Most existing approaches rely on supervised fine-tuning~\cite{ho2023large, magister2023teaching, pmlr-v202-fu23d} and  leverages advanced LLMs such as the GPT series~\cite{radford2019language, brown2020language, achiam2023gpt} as the guidance to generate high-quality data ~\cite{josifoski2023exploiting,li2023textbooks}. Symbolic Chain-of-Thought Distillation~\citep{li2023symbolic} introduced a step-by-step reasoning framework, highlighting the potential of distilling complex reasoning processes, while FLD~\citep{morishita2023learning} focused on logical deductive reasoning.
More recently, \citet{guo2025deepseek} proposed distilling multiple small models via supervised fine-tuning (SFT) over reasoning data generated by DeepSeek-R1. However, most methods treat LLMs merely as sources of reasoning chains, optimizing student models exclusively through supervised fine-tuning. Additionally, some works have explored reinforcement learning to further enhance reasoning capabilities; for instance, MARIO~\cite{ramnath2023tailoring} employs multiple external reward models to improve self-rationalization.
In contrast, our work takes a different approach by leveraging LLMs not only for response generation but also for extracting reward signals, enabling a more comprehensive distillation process.

\textbf{Reinforcement learning from External Feedback.} Following the success of Reinforcement Learning from Human Feedback (RLHF)~\citep{ziegler2019fine} in enabling the widespread application of large language models, researchers have increasingly explored ways to reduce human involvement in training through Reinforcement Learning from AI Feedback (RLAIF).
As a pioneer in RLAIF, Constitutional AI~\citep{bai2022constitutional} has demonstrated improved performance in tasks such as summarization, helpful dialogue generation, and harmless dialogue generation. \citet{Lee2023RLAIFVR} further showed that RLAIF can achieve performance comparable to or even surpassing RLHF, as evaluated by human judges. 
% Moreover, RLAIF has been shown to outperform a supervised fine-tuning baseline, even when the LLM preference labeler is the same size as the policy model.
SPIN~\cite{chen2024self} eliminates the need for explicit reward models by adopting an iterative DPO-like framework, where human-labeled winning responses are paired with the previous iteration’s generations as losing responses. However, those method do not focus on training smaller models through knowledge distillation.

% \textbf{LLM-as-a-Judge} Using LLM-as-a-Judge prompting to evaluate language models has become a standard approach~\cite{zheng2023judging, gu2024survey}[Dubois et al., 2023, Li et al., 2023, Fernandes et al., 2023, Bai et al., 2023, Saha et al., 2023], and is being used to train reward models or curate data as well, as described above [Lee et al., 2023, Chen et al., 2024a, Li et al., 2024]. While some works such as Kim et al. [2023] create training data to train an LLM to perform well as a judge, to our knowledge it is not common to combine this training with general instruction following skills as in our work.

\vspace{-2pt}
\section{Methodology}
\vspace{-2pt}
In this section, we present our problem formulation and outline the proposed pipeline for distilling a small language model by distilling both teacher's responses and rewards.

Our objective is to explore a new approach to improving student models through reinforcement learning by distilling both the responses and reward signals from teacher LLMs. Formally, we begin with \textbf{only} a large teacher model ($\mathcal{T}$), a smaller student model ($\mathcal{S}$), and a query dataset ($\mathcal{D}$) containing questions $q^i$ for all $i \in |\mathcal{D}|$.
We focus on tasks where, given a query $q$, the goal is to generate a complete response $a$, which consists of a Chain-of-Thought reasoning path $\bm{c}$ followed by a final answer $y$, as shown in Table~\ref{tab:qa_example}

The overall pipeline, illustrated in Figure~\ref{fig:framework}, consists of \textbf{ Data Distillation} and \textbf{Reward Distillation}.

\vspace{-5pt}
\subsection{Data Distillation}
In this phase, we autonomously generate and label responses to create a robust dataset for training both the reward model and fine-tuning the student model, given the query dataset $\mathcal{D}$ and the teacher model $\mathcal{T}$.

\textbf{Teacher's Generation.}
To construct this dataset, we generate multiple responses for each query using a powerful teacher model, $\mathcal{T}$, under two different settings, controlled by temperature during generation: \textit{high-confidence }(low temperature) and \textit{low-confidence} (high temperature). The \textit{high-confidence set} is designed to mimic annotators with the highest reliability, providing pseudo-labels $y$ for the queries and identifying cases where the teacher fails to produce a valid response. In contrast, the \textit{low-confidence set} serves to enhance response diversity during the warm-up stage, capturing more varied reasoning paths $\bm{c}$. By sampling responses across these two confidence categories, we ensure both reliability in pseudo-labeling and diversity in reasoning styles, thereby creating a more comprehensive training dataset. The prompt templates used to query the teacher model are provided in Appendix~\ref{app:prompt}.

\textbf{Teacher's Evaluation -- Pseudo Final Answer Voting.}
Evaluating responses is challenging because teacher models are primarily trained for text generation rather than assessment. Inspired by \citet{xiong2023can}, we adopt a majority voting approach instead of directly prompting the teacher model to evaluate each answer. This method not only enhances reliability but also reduces the computational cost of querying the teacher model.  
To ensure more reliable evaluations, we generate responses using a relatively low-temperature setting, reducing the uncertainty in the answers. These low temperature are set to $0$, $0.1$, $0.2$ and generate multiple responses for each temperature. From these \textit{high-confidence responses}, we derive a pseudo-final answer $y^*$ through majority voting. Specifically, if a single final answer $y$ appears with a probability exceeding a predefined threshold $sh=0.7$ among the high-confidence outputs, we designate it as the pseudo-final answer for the query and label the corresponding responses $a$ as correct; otherwise, they are marked as incorrect. If no single answer meets the threshold, we classify the query as one that the teacher model fails to answer, forming a set $\mathcal{D}_{\text{fail}}$.  
Finally, we propagate the pseudo-final answer to the responses in \textit{low-confidence response set}, assigning answers as either positive or negative accordingly.

\vspace{-2pt}
\subsection{Warm-Up via SFT} 
\vspace{-2pt}
% We first perform data distillation through supervised fine-tuning. We collect all teacher-generated responses that have been labeled as correct into the dataset for student warm-up,
We begin by performing data distillation through supervised fine-tuning (SFT). Specifically, we compile all teacher-generated responses that have been aligned with the pseudo final answer (labeled as correct) and incorporate them into the dataset for student warm up,
\begin{equation}
\small
    \mathcal{D}_{\text{sft}} = \{(q, a)\mid \text{extract}(a) = y^*\}, \, y^* \text{ is the pseudo label of } q.
\end{equation}
We then train the student model for two epochs. Notably, queries without a reliable pseudo-label—\emph{i.e.}, those that the teacher model fails to solve—are excluded from the SFT warm-up process. This ensures that the student model learns from diverse high-quality responses, forming a solid foundation before transitioning to reinforcement learning.  

The training objective in this stage is to minimize the negative log-likelihood (NLL) loss:
\begin{equation}
\small
   \mathcal{L}(\theta) = - \mathbb{E}_{(q, a)\sim \mathcal{D}_{\text{sft}}} \log P(a_k | a_{<k}, q, \theta),
\end{equation}
where $k$ denotes the index of words in the response $a$, including the final answer $y$ for the given query $q$, and $\theta$ represents the parameters of the student model. We obtain a student model $\mathcal{S}_{\text{warmup}}$ after warm up phase.

\subsection{Reward Distillation I: Reward Model Learning}
% Reward Model Learning}
\label{sec:rew}
Below, we describe how we distill the teacher model's evaluation capability through reward model learning.

We propose a self-supervised schema that leverages the inherent structure of NLP tasks. This schema evaluates both \emph{teacher-generated} and \emph{student-generated} responses, systematically collecting, labeling, and filtering them through structured constraints to construct reward training data.

A straightforward and efficient approach is to train a reward model to distinguish whether the teacher’s answers are correct. However, this method can introduce significant distribution shift due to differences in generation patterns between the teacher and student models. To mitigate this issue, we primarily train the reward model on the student model’s responses while also incorporating the teacher’s responses for additional guidance.

\textbf{Student's Generation \& Self-supervised Evaluation.}  
% To begin with, we collect multiple responses from the student model for each query $q$ and require self-supervised evaluation by aligning the student’s answers with the pseudo labels. During student's generation, we use a temperature of $0.7$ to ensure balanced diversity and confidence. This process forms a dataset:
Using a temperature of \(0.7\), we generate multiple responses from the student model for each query \( q \), ensuring less confidence but diverse outputs. We then align these student responses with pseudo labels derived from the teacher’s self-evaluations, constructing the dataset: 
\begin{equation}
\small
    \mathcal{D}_{\mathcal{S}} := \{(q, a^{\mathcal{S}}, y^*) \mid \forall q \notin \mathcal{D}_{\text{fail}}, \exists a \in a^{\mathcal{S}}, y \neq y^*\}
\end{equation}
where $a^{\mathcal{S}}$ represents the set of student-generated answers, and $y = \text{extract}(a)$ is the extracted final answer from a student response $a$, which is compared against the pseudo-label $y^*$. Queries without a reliable pseudo-label, i.e., those in $\mathcal{D}_{\text{fail}}$, are excluded from this dataset. 
% \paragraph{Applying the Constraints to Teacher \& Student Responses.}
% These constraints are applied uniformly to both \emph{teacher-generated} and \emph{student-generated} responses:
% \begin{enumerate}[leftmargin=1em]
%     \item \textbf{Teacher outputs} are gathered under both high and low temperature settings, with pseudo-final answers derived from majority voting (\S\ref{sec:teacher_eval}).
%     \item \textbf{Student outputs} come from the student model (after SFT warm-up) at a moderate temperature, ensuring diverse yet coherent responses.
% \end{enumerate}
% Each response is then labeled \emph{positive} or \emph{negative} based on how many of the above constraints it satisfies (e.g., correct final answer, coherent reasoning, extractable solution, and semantic alignment). These labels form the basis for training our reward model, as detailed in \S\ref{sec:reward_data} and \S\ref{sec:reward_objective}.

\textbf{Constructing Training Data for Reward Model Learning.}  
To determine which queries are used for reward model learning, we start with the dataset $\mathcal{D}_{\mathcal{S}}$. A query $q$ is in the dataset $\mathcal{D}_{\mathcal{S}}$ for reward model learning if at least one of the warm-up model’s final answers, extracted from the response $a^{\mathcal{S}}$, differs from its pseudo-label $y^*$.  

For each selected query \( q \) in $\mathcal{D}_{\mathcal{S}}$, we uniformly sample an equal number of:  
\begin{itemize}[nosep]
    \item positively labeled student responses $a^{\mathcal{S}^+}$,
    \item negatively labeled student responses $a^{\mathcal{S}^-}$,
    \item positively labeled teacher responses $a^{\mathcal{T}^+}$, where $a^{\mathcal{T}}$ represents all teacher-generated responses.

\end{itemize}  

This process constructs the dataset for reward model learning, denoted as:  
\begin{equation}
\small
    \mathcal{D}_{\mathcal{R}}:=\{(q, (a^{\mathcal{S}^+}, y^{\mathcal{S}^+}), (a^{\mathcal{S}^-}, y^{\mathcal{S}^-}), (a^{\mathcal{T}^+}, y^{\mathcal{T}^+}))\},
    \nonumber
\end{equation}
where $y^{\mathcal{S}^+} = y^{\mathcal{T}^+} = y^*$ and $y^{\mathcal{S}^-} \neq y^*$.  

Our optimization objective comprises two key components: classification and preference modeling. Classification labels are directly derived from the teacher’s evaluations, while we additionally construct preference labels to guide the reward model in distinguishing both correctness and reasoning quality: for a given query $q$,  the student model’s positive responses receive higher rewards than its negative ones; the teacher’s positive responses are rewarded more highly than the student model’s positive responses. This design is motivated by the observation that a student model may follow an incorrect reasoning path yet arrive at the correct final answer, whereas the teacher model typically provides a more concise and reliable reasoning process. By structuring the dataset in this manner, the reward model is exposed to a diverse range of responses, enabling it to differentiate not only between correct and incorrect answers but also to prefer responses that demonstrate more accurate reasoning.

\textbf{Reward Model Learning Objective.}  
Given the constructed dataset $\mathcal{D}_{\mathcal{R}}$, we define the learning objective for our reward model, $\mathcal{R}$. Specifically, $\mathcal{R}$ takes a question-answer pair $(q, a)$ as input and outputs a scalar reward $r$ that distinguishes between high-quality and low-quality responses. The reward model is initialized from the warm-up student model, $\mathcal{S}_{\text{warmup}}$, and its training loss comprises two components: a binary cross-entropy (BCE) classification loss and a preference-based loss.

The classification loss ensures that the reward model correctly assigns higher scores to correct responses and lower scores to incorrect ones:  
\begin{equation}
\begin{aligned}
\small
    \mathcal{L}_{\text{cls}}(\mathcal{R}) = & - \mathbb{E}_{(q, a, y, y^*) \sim \mathcal{D}_{\mathcal{R}}} 
    \Bigl[ 
        \mathbbm{1}(y = y^*) \log \sigma(\mathcal{R}(q, a)) \\
        & + (1 - \mathbbm{1}(y = y^*)) \log (1 - \sigma(\mathcal{R}(q, a))) 
    \Bigr],
\end{aligned}
\label{eq:reward_classification}
\end{equation}
where $\sigma$, $\mathbbm{1}$ is the sigmoid and indicator function.  

Additionally, we introduce a preference-based loss to enforce ranking constraints, ensuring that:  
1) Positive student responses are preferred over negative ones.  
2) Teacher-generated positive responses are preferred over student-generated positive responses.  Therefore, the preference-based loss is formulated as:  
\begin{equation}
\resizebox{0.98\linewidth}{!}{$
\begin{aligned}
\small
   \mathcal{L}_{\text{pref}}(\mathcal{R})= &- \mathbb{E}_{q, a^{\mathcal{S}^+}, a^{\mathcal{T}^+} \sim \mathcal{D}_{\mathcal{R}}} 
    \Bigl[
        \log \sigma\bigl(\mathcal{R}(q, a^{\mathcal{T}^+}) - \mathcal{R}(q, a^{\mathcal{S}^+})\bigr)
    \Bigr] \\
    & - \mathbb{E}_{q, a^{\mathcal{S}^+}, a^{\mathcal{S}^-} \sim \mathcal{D}_{\mathcal{R}}} 
    \Bigl[
        \log \sigma\Bigl(\mathcal{R}\bigl(q, a^{\mathcal{S}^+}\bigr) - \mathcal{R}\bigl(q, a^{\mathcal{S}^-}\bigr)\Bigr)
    \Bigr],
\end{aligned}
\label{eq:reward_preference}
$}
\end{equation}
where \(\mathcal{R}(q, a)\) represents the predicted reward for response $a$ given query $q$.  

Finally, the objective function for reward model training is,   
\begin{equation}
\small
        \mathcal{L} = \mathcal{L}_{\text{pref}} + \lambda \mathcal{L}_{\text{cls}}.
\label{eq:reward_total}        
\end{equation}

\subsection{Reward Distillation II: Optimization through Reinforcement Learning}

With the trained reward model $\mathcal{R}$, we further refine the student model $\mathcal{S}$ after the warm-up phase using reinforcement learning (RL).  

\textbf{Data.}  
The training data for RL is drawn from $\mathcal{D}_{\mathcal{R}}$ and $\mathcal{D}_{\text{fail}}$, where $\mathcal{D}_{\mathcal{R}}$ consists of queries that the student failed to address, and $\mathcal{D}_{\text{fail}}$ contains queries that the teacher was unable to solve.

\textbf{Reward Design.}  
The quality of the reward model has a significant impact on the optimization process and performance of PPO. To ensure that it captures multiple aspects of response quality while remaining computationally efficient during Reinforcement Learning (RL), we augment our trained reward model with the following design:  

\begin{enumerate}[nosep]
    \item \textbf{Answer Existence and Extractability.}  
    If the model fails to provide an extractable answer, we assign a reward of $-5$ and terminate the evaluation for this query. 
    \item \textbf{Reward Model Score.}  
    We use the predicted reward from the trained reward model $\mathcal{R}$. Ideally, a positive reward value indicates correctness, while a higher reward reflects a better response.  

    \item \textbf{Consistency Check with the Pseudo Final Answer in $\mathcal{D}_{\mathcal{R}}$.}  
    Finally, we compare the extracted answer with the pseudo-final answer $y^*$, if applicable. If the extracted value does not match the pseudo-label, we adjust the reward using $\min(r, 0)$.  
\end{enumerate}  
Therefore, the final reward for a response $a$ for question $q$ is defined as:  
\begin{equation}
\small
      \Tilde{\mathcal{R}}(q, a) =
    \begin{cases}
      -5, & \text{if no extractable answer is found};\\
      \min(r, 0), & \text{if } \hat{y} \neq y^*;\\
      r, & \text{if } \hat{y} = y^*,
    \end{cases}  
    \label{eq:reward_ppo}
\end{equation}
where $r=\mathcal{R}(q, a)$ is the predicted reward from the reward model.  We use the augmented $\Tilde{\mathcal{R}}(q, a) $ in the further reinforcement learning.

This design ensures that the student model is discouraged from producing incomplete responses while also allowing the reward model’s output to be further refined by aligning it with the pseudo-labeled final answer when applicable.

\textbf{Optimization with PPO.}  
We employ Proximal Policy Optimization (PPO)~\cite{schulman2017proximal} in RL to refine the student model $\mathcal{S}_{\text{warmup}}$ under the supervision of reward signals defined in Eq.~\ref{eq:reward_total}. Let $\theta$ denote the parameters of  $\mathcal{S}_{\text{warmup}}$.  The loss function for optimizing $\theta$ is,
\begin{equation}
\resizebox{0.9\linewidth}{!}{$
\begin{aligned}
\small
 \mathcal{L}_{\text{PPO}}= & \mathbb{E}_{(q,a)\sim \mathcal{D}_{\mathcal{R}}\cup\mathcal{D}_{\text{fail}} } \left[ 
    \min ( m_t(\theta) \hat{A}_t,\, \right.\\ &
        \left. \text{clip}(m_t(\theta), 
         1 - \epsilon, 1 + \epsilon) \hat{A}_t 
    ) \right] \\
& \quad - \lambda\, \mathbb{E} \left [
    \text{KL}(
        \pi_{\theta}(\cdot \mid  a_{<t}; q) 
        \,\|\, 
        \pi_{\theta_{\text{old}}}(\cdot \mid  a_{<t}; q))
\right]
% \\
% & m_t(\theta) = \frac{\pi_{\theta}(a_t\mid a_{<t}, q)}{\pi_{\theta_{\text{old}}}(a_t\mid a_{<t}; q)},
\end{aligned}
$}
\end{equation}
where \( m_t(\theta) = \frac{\pi_{\theta}(a_t\mid a_{<t}, q)}{\pi_{\theta_{\text{old}}}(a_t\mid a_{<t}; q)} \) represents the probability ratio between the updated policy \(\pi_{\theta}\) and the previous policy \(\pi_{\theta_{\text{old}}}\). The term \( \hat{A}_t \) is the estimated advantage function at time step \( t \), while \( \epsilon \) is the clipping threshold that constrains policy updates. The coefficient \( \lambda \) controls the penalty term, and  
$\text{KL}\left[ \pi_{\theta}(\cdot|a_{<t}; q) \,\Big\|\, \pi_{\theta_{\text{old}}}(\cdot|a_{<t}; q) \right]$ is the Kullback-Leibler (KL) divergence between the current and previous policies, ensuring stable updates.  

\begin{table*}[!ht]
\centering
\small
\caption{Evaluation results across different teachers, students and datasets. 
The best scores for each model size are boldfaced, and the scores where the student model outperforms the teacher are marked in \textcolor{blue}{blue}.}
\begin{tabular}{llcccc}
\toprule
\textbf{Teacher} & \textbf{Student} & \textbf{Method} & \textbf{GSM8K} & \textbf{GSM-Plus} & \textbf{MMLU-Pro}\\
\midrule
\midrule
& & Llama3-70B (Teacher) & 93.18\% & 83.24\% & 56.85\% \\
\midrule
\multirow{10}{*}{Llama3-70B}
& \multirow{5}{*}{Llama3-1B} & ICL & 41.55\% & 28.92\% & 15.97\% \\
& & SFT & 61.03\% & 39.95\% & 31.00\%\\
& & Ours w/o R & 61.41\% & 40.38\% & 32.83\%\\
& & Ours w/o D & 61.25 \% & \textbf{42.86 \%} & 26.16\% \\
& & Ours & \textbf{64.06\%} & 41.86\%& \textbf{35.27\%}\\
\cline{2-6}
& \multirow{5}{*}{Llama3-3B} & ICL & 72.48\% & 45.48\% & 29.62\% \\
& & SFT & 80.74\% & 61.76\% & 41.54\%  \\
& & Ours w/o R & 80.89\% & 62.24\% &  42.10\%\\
& & Ours w/o D & 80.29 \% & \textbf{63.76\% }& 35.30\% \\
& & Ours & \textbf{82.64\%} & 63.05\% & \textbf{44.50\%}\\
\midrule
\midrule
& & Llama3-8B (Teacher) & 80.97\% & 68.19\% & 39.88\% \\
\midrule
\multirow{10}{*}{Llama3-8B} 
& \multirow{5}{*}{Llama3-1B} & ICL & 41.55\% & 28.92\% & 15.97\% \\
& & SFT & 62.85\% & 42.23\% & 22.62\% \\
& & Ours w/o R & 63.08\% & \textbf{43.14\%} & 23.85\%\\
& & Ours w/o D & 60.12\% & 40.95\% & 26.44\% \\
& & Ours & \textbf{66.03\%} & 43.00\% & \textbf{32.00\%} \\
\cline{2-6}
& \multirow{5}{*}{Llama3-3B} & ICL & 72.48\% & 45.48\% & 29.62\%\\
& & SFT & 79.75\% & 62.85\% & 31.78\% \\
& & Ours w/o R & 79.91\% & 62.86\%  & 34.50\% \\
& & Ours w/o D & 80.36\% & 64.19\%  & 36.08\% \\
& & Ours & $\textcolor{blue}{\textbf{83.02\%}}$  & \textbf{64.24\%} & $\textcolor{blue}{\textbf{40.02\%}}$\\
% \midrule
\midrule
\bottomrule
\label{tab:main_results}
\vspace{-15pt}
\end{tabular}
\end{table*}

\section{Experiments}
In this section, we evaluate the efficacy of our proposed method and conduct comprehensive ablation studies to assess its effectiveness and justification.
\subsection{Setup}
\textbf{Evaluation Datasets.}  
We evaluate our method on three widely used benchmarks: \textbf{GSM8K}~\citep{cobbe2021gsm8k}: A dataset of grade-school math problems designed to assess models' problem-solving and reasoning capabilities;  \textbf{GSM-Plus}~\citep{li2024gsm}: A variant of GSM8K that introduces various mathematical perturbations to test generalization. We train student models on GSM8K and directly evaluate them on GSM-Plus; \textbf{MMLU-Pro}~\citep{wang2024mmlu}: A professional-level multi-task benchmark covering a wide range of knowledge domains, including humanities, sciences, and engineering.
\textbf{Data Split:} Details of the dataset splits are provided in Appendix~\ref{app:data_split}. During training, we select the optimal checkpoints based on the student model’s performance on the validation set.  

\textbf{Evaluation Metrics.}  
We report accuracy as the primary evaluation metric. To ensure reproducibility, we set the generation temperature to $0$ during inference.

\textbf{Implementation Details.}  
For our experiments, we implement the proposed method using a combination of different large language models (LLMs) as teachers, smaller models as students, and classifiers as reward models. \textbf{Teachers:}  We employ Llama3-70B\footnote{\href{https://huggingface.co/meta-llama/Meta-Llama-3-70B-Instruct}{https://huggingface.co/meta-llama/Meta-Llama-3-70B-Instruct}} and Llama3-8B\footnote{\href{https://huggingface.co/meta-llama/Llama-3.1-8B}{https://huggingface.co/meta-llama/Llama-3.1-8B}}~\citep{dubey2024llama}. As shown in Table~\ref{tab:main_results}, Llama3-70B achieves 93.18\% accuracy on GSM8K and 56.85\% on MMLU-PRO, while Llama3-8B achieves 80.97\% on GSM8K and 39.88\% on MMLU-PRO. This allows us to evaluate our method across teacher models with varying levels of knowledge. \textbf{Students:}  
    We use Llama3-1B\footnote{\href{https://huggingface.co/meta-llama/Llama-3.2-1B}{https://huggingface.co/meta-llama/Llama-3.2-1B}} and Llama3-3B\footnote{\href{https://huggingface.co/meta-llama/Llama-3.2-3B}{https://huggingface.co/meta-llama/Llama-3.2-3B}}~\citep{dubey2024llama}. These smaller models serve as students, learning and improving through knowledge distilled from the teacher models.  \textbf{Reward Models:}  
    The reward model is based on the student model with an additional linear predictor on top of its final embedding layer, producing a scalar reward estimate. It is fine-tuned in a single-head setting to evaluate responses generated by the student model, taking both the question and answer as inputs.  \textbf{Data collection}, each query is used to prompt the teacher model five times at temperatures 0, 0.1, 0.2, and 0.3 to collect high-confidence responses. For low-confidence responses, the teacher is prompted five times at temperatures of $0.4$, $0.5$, $0.6$, $0.7$, $0.8$, $0.9$ and $1.0$.  For reward model learning, we infer the student model (after warm-up) 30 times per question. We set $\lambda$ in Eq.~\ref{eq:reward_total} to 0.5.  For more hyperparameters, please refer to the appendix~\ref{app:hyperparameters}.

\textbf{Evaluated Methods.}
To assess the efficacy of our approach, we report the following baselines and ablation variants:  \textbf{In-Context Learning (ICL):} The student model utilizes in-context learning without any fine-tuning.  \textbf{Supervised Fine-Tuning (SFT):} The student model is trained via Supervised Fine-Tuning (SFT) on the teacher's responses. \textbf{Teacher LLM:} We report the performance of teacher LLMs (Llama3-70B or Llama3-8B) to evaluate whether the student can outperform the teacher. \textbf{Ours:} Our full method. The student model is trained with reinforcement learning (RL) following a warm-up phase using the teacher's responses.  \textbf{Ours w/o Data (Ours w/o D):} The student model is trained using RL in the same manner as \textbf{Ours}, but without the SFT warm-up phase utilizing the teacher's responses. \textbf{Ours w/o Reward (Ours w/o R):} The student model is trained using SFT on a dataset generated by the teacher, with the teacher's self-evaluation employed to filter high-quality responses for training.

\subsection{Main Results}
In this subsection, we report the performance of teacher and student models of different sizes on three benchmark tasks. The experimental results are provided in Table~\ref{tab:main_results}. Note that for GSM-Plus, the models are trained on GSM8K and evaluated on GSM-Plus, demonstrating the ability to generalize beyond the direct training task. According to the experimental results, 
\textbf{our method consistently outperform other methods across varying teacher model capabilities.} Starting with the stronger teacher, Llama3-70B, compared with SFT, the 1B student’s accuracy are increased from 61.03\% to 64.06\% on GSM8K (+3.03\%) and from 31.00\% to 35.27\% on MMLU-Pro (+4.27\%); for a 3B student, our method achieves a +1.90\% improvement on GSM8K and +2.96\% on MMLU-Pro .  
When shifting to the less capable teacher, Llama3-8B, the improvement is more obvious: our 1B student still attains a +3.18\% gain on GSM8K and a substantial +9.38\% jump on MMLU-Pro (22.62\% to 32.00\%). The 3B student follows a similar trend, with +3.27\% on GSM8K and +8.24\% on MMLU-Pro (31.78\% to 40.02\%).  
These results highlight the robustness of our method, which consistently yields performance improvements across different student sizes and teacher capabilities. 
Notably, in certain configurations—shown in \textcolor{blue}{blue}—\textbf{our method even enable the students outperform their respective teachers.}  For instance, under Llama3-8B, the 3B student surpasses the teacher on GSM8K (83.02\% vs. 80.97\%) and on MMLU-Pro (40.02\% vs. 39.88\%), demonstrating the effectiveness of our distillation approach in transferring knowledge with both data and rewards.  Additionally, our method also provid a substantial boost to student models even under cross-task generalization (from GSM8K to GSM-Plus).
% Our method shows a notable improvement in performance on all datasets, with particularly strong results on GSM8K, demonstrating its effectiveness in tasks requiring reasoning.

\subsection{Ablation Study}  
To further analyze the rationale behind our method, we conduct ablation studies from various perspectives. Unless otherwise specified, these ablations are performed using the teacher model Llama3-70B, the student model Llama3-1B, and the GSM8K dataset.

\begin{figure}[!ht]
    \centering
    \includegraphics[width=0.9\linewidth]{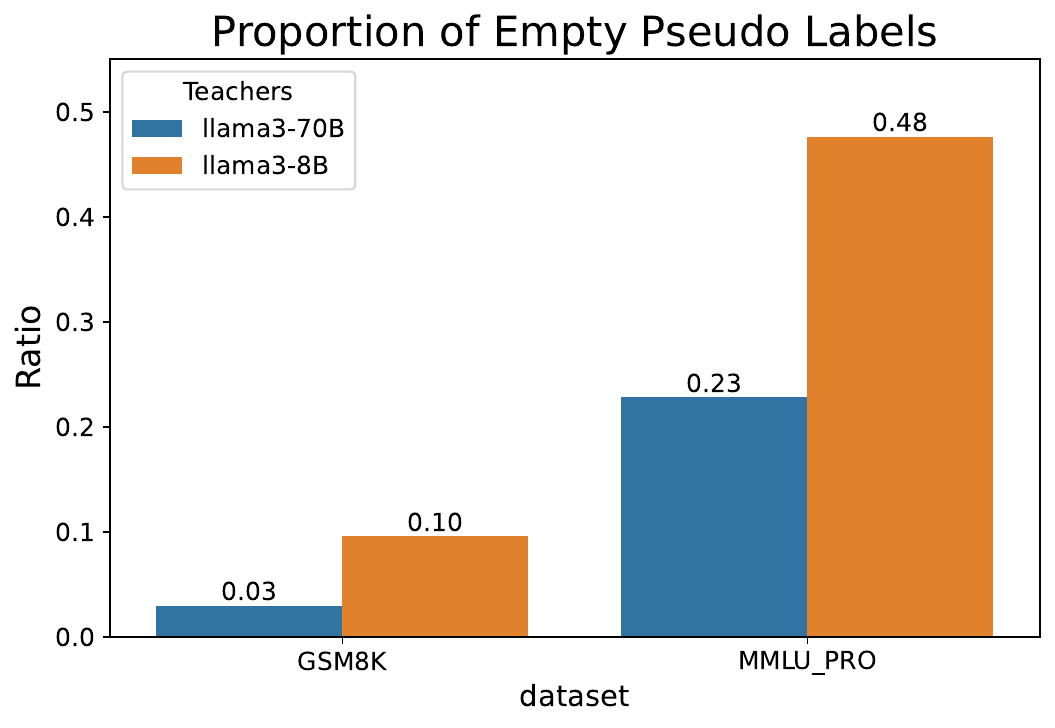}
    \vspace{-15pt}
    \caption{Visualization of the ratio of questions without reliable pseudo labels based on the teacher's evaluation.}
    \label{fig:teacher_fail}
    \vspace{-5pt}
\end{figure}

\begin{figure}[!ht]
    \centering
    \includegraphics[width=\linewidth]{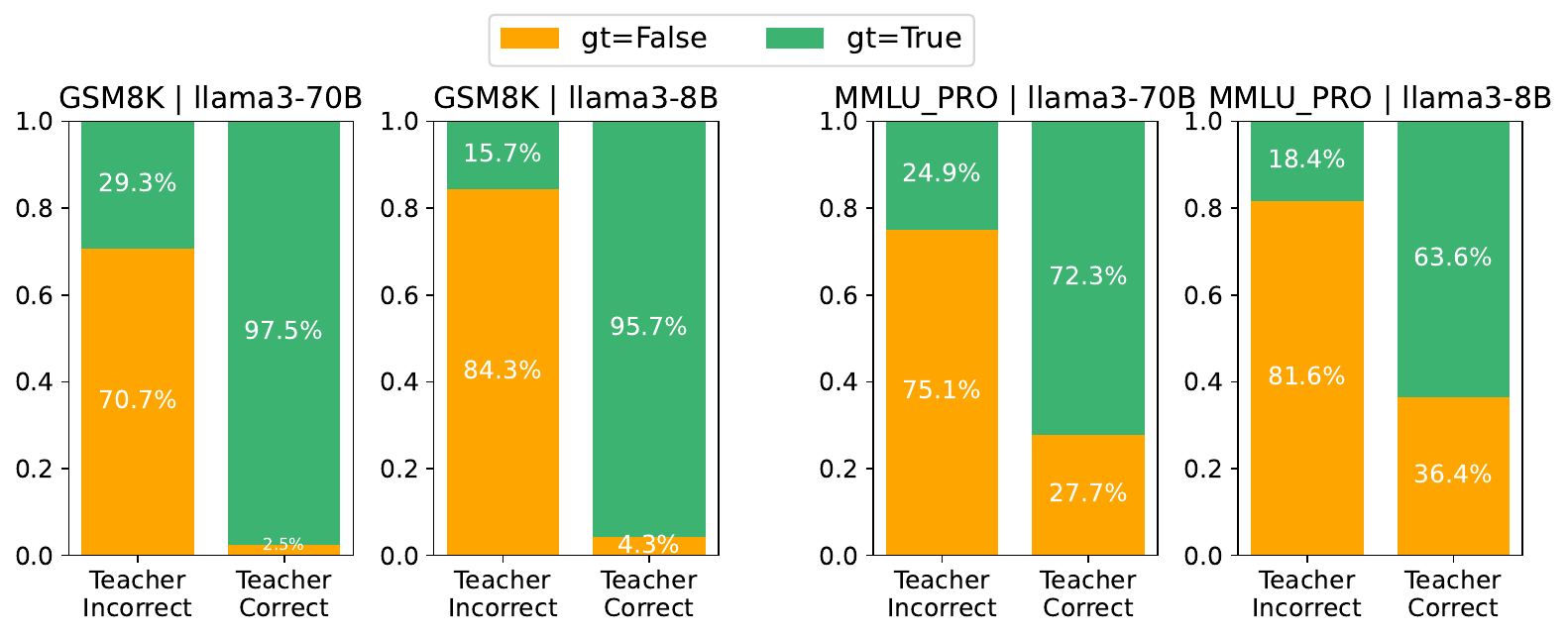}
    \vspace{-15pt}
    \caption{Visualization on teacher's evaluation (pseudo labels) across different datasets and teachers. For each subfigure, we represent the samples which are regarded as incorrect by teacher in the left and correct ones in the right. We represent the judgement by ground truth correct in yellow and incorrect in green.}
    \label{fig:noisy_data}
    \vspace{-5pt}
\end{figure}

\textbf{Visualization of the Teacher's Evaluation Capacity.}  
We visualize the teacher model's evaluation capability in Figure~\ref{fig:teacher_fail} and Figure~\ref{fig:noisy_data}.  Figure~\ref{fig:teacher_fail} presents the proportion of questions for which the teacher fails to reliably generate a pseudo-final answer. The results indicate that while teacher models exhibit near-perfect evaluation on GSM8K, their performance drops significantly on MMLU-PRO. This issue is particularly evident with Llama3-8B, where nearly half of the questions fail to obtain a reliable pseudo-final answer.  
Furthermore, for questions with pseudo-final answers, we visualize the proportion of samples that the teacher model classifies as correct. As shown in Figure~\ref{fig:noisy_data}, the inaccuracy of the teacher's evaluation increases from Llama3-70B to Llama3-8B and from GSM8K to MMLU-PRO. In summary, evaluation capacity is weakest on MMLU-PRO, especially for Llama3-8B. Interestingly, as shown in Table~\ref{tab:main_results}, despite the most challenging setting—MMLU-PRO with Llama3-8B as the teacher—our method achieves the largest performance improvement, highlighting its robustness in scenarios with weaker teacher supervision. The probable reason behind this is that the student gains the least knowledge from warm-up, therefore leaving much room for improvement.

\textbf{Performance of Reward Model.} We report the performance of reward model distinguish the positive and negative student responses (\textbf{Pos/Neg}) and the positive student and teacher responses (\textbf{Pos/Pos}), shown in Table~\ref{tab:rm_acc}.
\begin{table}[t]
    \centering
    \caption{Evaluation results with reward model performance.}
\small
    \begin{tabular}{llcccc}
        \toprule
        \multirow{2}{*}{$\mathcal{T}$} & \multirow{2}{*}{$\mathcal{S}$} & \multicolumn{2}{c}{GSM8K} & \multicolumn{2}{c}{MMLU-Pro} \\
        \cmidrule(lr){3-4} \cmidrule(lr){5-6}
        & & Pos/Neg & Pos/Pos & Pos/Neg & Pos/Pos \\
        \midrule
        \multirow{2}{*}{70B} & 1B &   0.7042 &  0.7223 &  0.6638 & 0.7752 \\
        & 3B & 0.7081 & 0.7034  &   0.6670 & 0.9027 \\
        \midrule
        \multirow{2}{*}{8B} & 1B & 0.6573 & 0.8829 & 0.7417 & 0.8510  \\
        & 3B &  0.6819 & 0.9382 & 0.7609 & 0.8798 \\
        \bottomrule
    \end{tabular}
    \label{tab:rm_acc}
    \vspace{-8pt}
\end{table}

%\textbf{Data choice for reward model learning.}
%We conduct ablation studies on the choice of data for reward model learning. 
\textbf{Comparison of Data Sources for Reward Model Training.}
We conduct ablation studies to assess the impact of different data sources for training the reward model on policy learning. In the ablation setup, we train a classification-based reward model using the loss function \( L_{\text{cls}} \) defined in Eq.~\ref{eq:reward_classification}, utilizing only the teacher's responses. As presented in Table~\ref{tab:rm_data}, when the reward model is trained exclusively on the teacher's responses, the student model exhibits a performance increase of \( +1.44\% \). In contrast, when the reward model is trained using the student's own responses, the student model achieves a larger performance gain of \( +2.65\% \). This difference can be attributed to the fact that reward models trained on the teacher's responses experience a greater distribution shift during policy learning, which adversely affects performance.

\begin{table}[!ht]
    \centering
    \vspace{-10pt}
    \caption{Impact of Reward Model Data Source on Student's Performance.}
    \small
    \begin{tabular}{c|cc}
    \hline
    w/o Reward & from Teacher & Ours (from Student)\\ \hline
    61.41\% & 62.85\% & 64.06\%  \\ \hline
    \end{tabular}
    \label{tab:rm_data}
    \vspace{-5pt}
\end{table}

% \textbf{Comparison with }
% We evaluate an ablated version of our full method by removing reward distillation and only perform SFT on filtered teacher's answers, referred to as \textbf{Ours w/o Reward}, as shown in Table~\ref{tab:main_results}.  

\begin{figure}[!ht]
    \centering
    \includegraphics[width=0.8\linewidth]{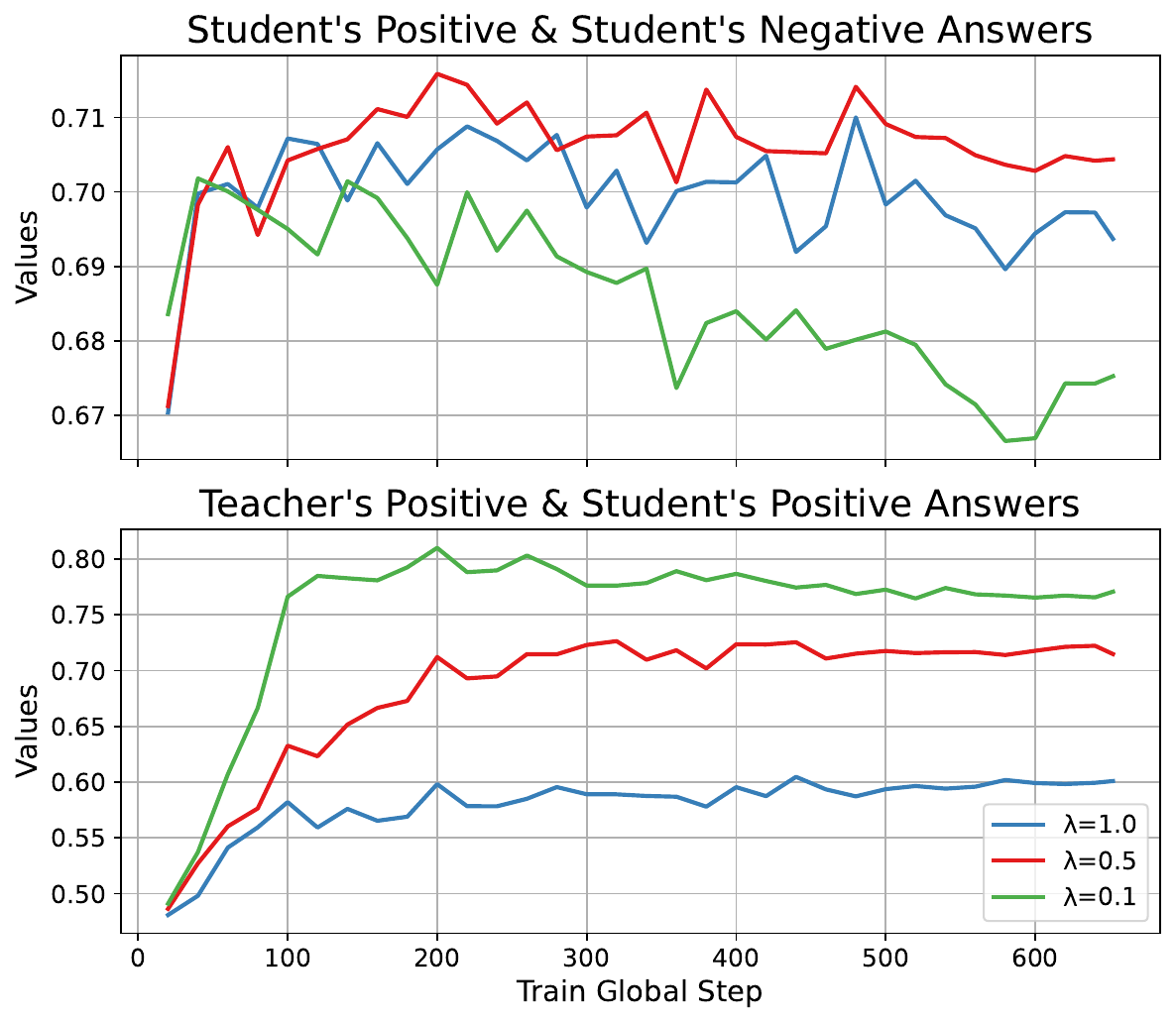}
    \vspace{-10pt}
    \caption{Ablation on $\lambda$ in reward model learning. We keep the evaluation set the same.}
    % \vspace{-5pt}
    \label{fig:ablation_lambda}
\end{figure}

\textbf{Impact of Hyperprameter $\lambda$ in Eq.~\ref{eq:reward_total} on Reward Model's Performance.}
We conduct ablation study on the hyperparameter $\lambda$ to see the impact on reward model's capacity of distinguishing the positive and negative answers. As shown in Figure~\ref{fig:ablation_lambda}, $\lambda=0.5$ enables a more stable learning process. 

\begin{figure}
    \centering
    \includegraphics[width=0.8\linewidth]{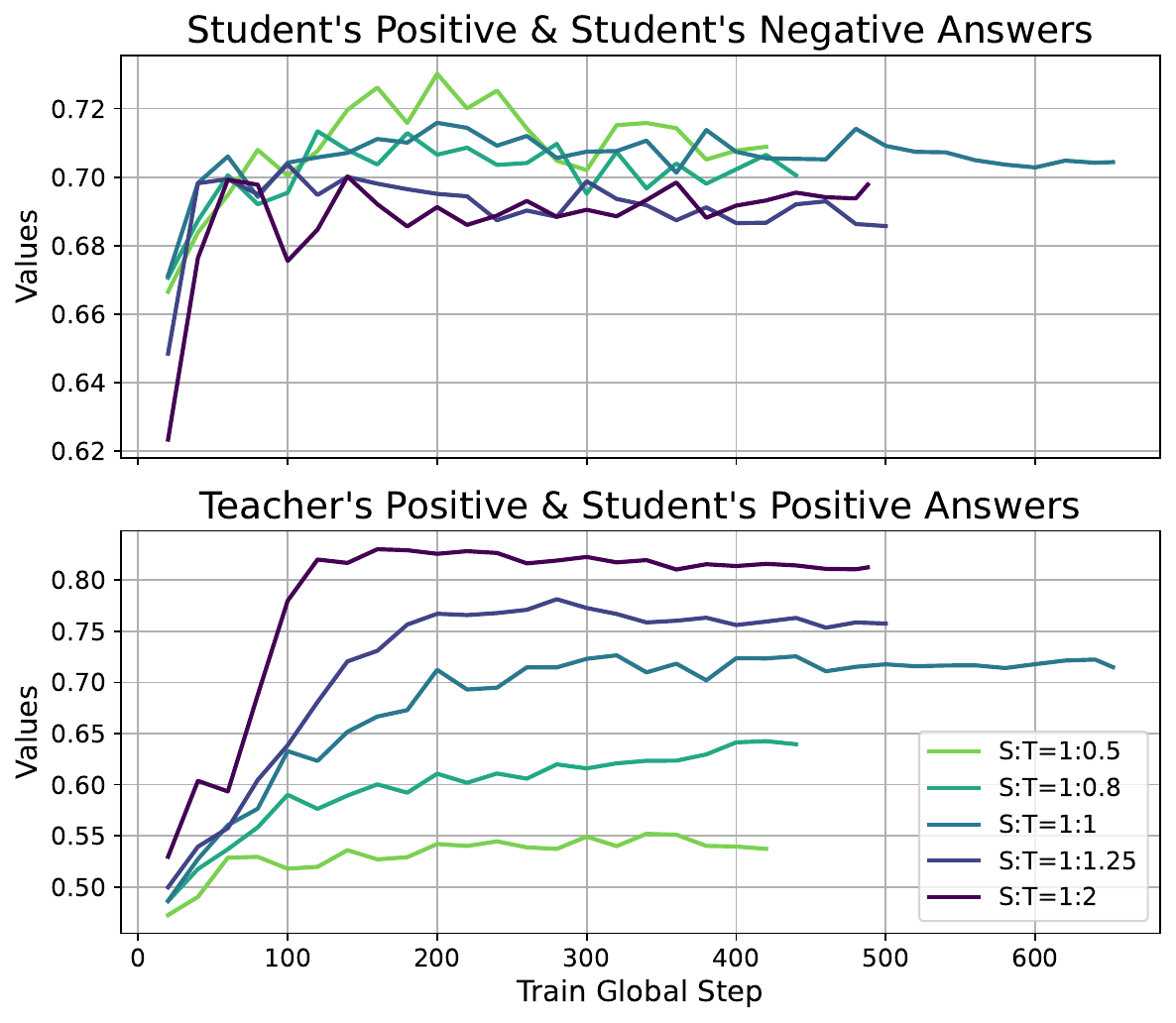}
    \vspace{-10pt}
    \caption{Variation in the ratio of teacher and student responses during reward model training. Darker colors indicate a larger proportion of teacher responses within the total set of responses. We evaluate each in the same validation set.}
    \label{fig:dataratio}
    \vspace{-5pt}
\end{figure}

\textbf{Impact of Data Ratio between Teacher's and Student's Answers on Reward Model Training.}
We adjust the data ratio of teacher and student responses during reward model training. As illustrated in Figure~\ref{fig:dataratio}, the data ratio has little effect on the reward model's ability to distinguish between good and bad responses but significantly influences its capacity to identify superior positive responses from both the teacher and the student.
% \textbf{Discussion on Interactive Improvement on Reward Model.}
% In order to investigate if there still room for improving student though further refining the reward model using the optimized student after PPO, we train a new reward model in the same way as Sec~\ref{sec:rew} and report the performance in Table~\ref{tab:interactive_ppo}.

% \begin{table}[t]
%     \centering
%     \caption{Evaluation results of interactively training reward model using student after RL on GSM8K.}
%     \begin{tabular}{llcc}
%         \toprule
%         Teacher & Student & $1^{st}$ PPO & $2^{st}$ PPO \\
%         \midrule
%         \multirow{2}{*}{70B} & 1B & 64.06\% &  63.76\% \\
%         & 3B & 82.64\% & 82.33\% \\
%         \midrule
%         \multirow{2}{*}{8B} & 1B & 66.03\% & 65.95\% \\
%         & 3B & 83.02\% &  83.50\% \\
%         \bottomrule
%     \end{tabular}
%     \label{tab:interactive_ppo}
% \end{table}

\begin{figure}[!ht]
    \centering
    % \vspace{-5pt}
    \includegraphics[width=0.7\linewidth]{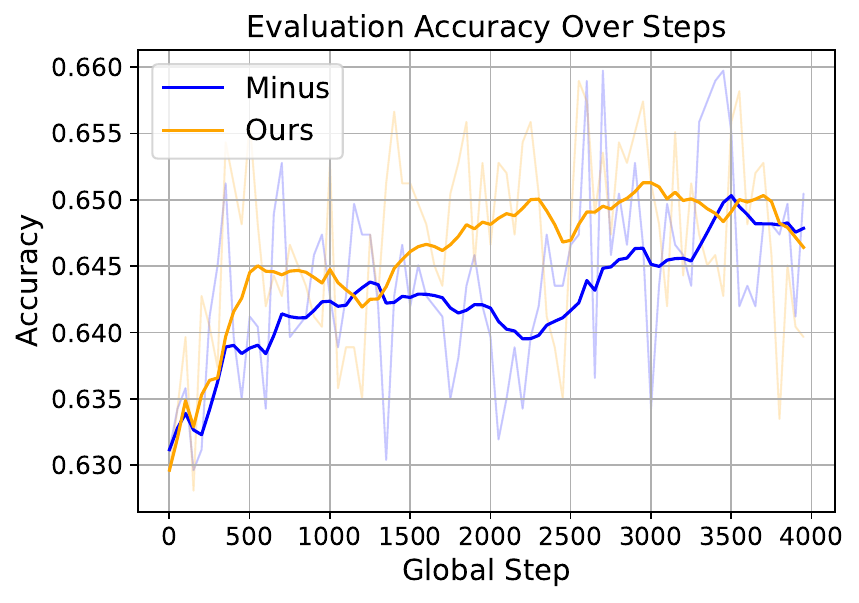}
    \vspace{-8pt}
    \caption{Different choice for correcting the reward model's prediction during PPO.}
    \label{fig:reward_choice}
    \vspace{-12pt}
\end{figure}

\textbf{Ablation on Design of Correctness Validation in Eq~\ref{eq:reward_ppo}.}
We conduct an ablation study on methods to correct the reward model's predictions. In our full method, we apply $\min(r, 0)$ when the extracted answer $y$ differs from the pseudo label $y^*$. Additionally, we evaluate an ablation variant that applies $r - 1$ when $y \neq y^*$ (\textbf{Minus}), as shown in Table~\ref{fig:reward_choice}.

\textbf{Comparison with Reward-Only Distillation and Data-Only Distillation.}
We evaluate two ablation versions of our full method by removing data distillation (warm-up), referred to as \textbf{Ours w/o Data}, and by removing reward distillation while only performing SFT on the filtered answers from the teacher, referred to as \textbf{Ours w/o Reward}, as shown in Table~\ref{tab:main_results}. Whether distilling only reward signals or only data from the teacher model can significantly improve the student model. However, \textbf{Ours} outperforms them by distilling both data and reward simultaneously.

\section{Conclusion}
In summary, we propose a novel distillation framework that leverages both teacher-generated outputs and self-supervised reward signals to train a student model. By introducing reinforcement learning on top of SFT-based data distillation, this approach effectively sidesteps the biases of direct teacher evaluations and addresses the mismatch between the student model’s inputs and the reward model in later training stages. Experimental results on GSM8K and MMLU-PRO demonstrate that this method not only outperforms purely SFT-based distillation strategies but also enables the student model to exceed the teacher’s performance in certain metrics. Our work highlights the untapped potential of exploiting teacher LLMs' reward signals and offer a new, scalable paradigm for distilling large language models when reliable direct evaluation signals are absent.
\newpage

\section*{Impact Statement}
Our work seeks to contribute to the advancement of machine learning by enhancing the efficiency and scalability of knowledge distillation for large language models. By incorporating both generative outputs and self-supervised reward signals, our approach minimizes dependence on explicit teacher evaluations and human-labeled data. While this can improve accessibility and efficiency in model training, it also introduces challenges related to bias propagation and the reliability of self-supervised reward modeling. We recognize these concerns and encourage further research to ensure the robustness and fairness of such methods in real-world applications.


\bibliography{example_paper}
\bibliographystyle{icml2025}

%%%%%%%%%%%%%%%%%%%%%%%%%%%%%%%%%%%%%%%%%%%%%%%%%%%%%%%%%%%%%%%%%%%%%%%%%%%%%%%
%%%%%%%%%%%%%%%%%%%%%%%%%%%%%%%%%%%%%%%%%%%%%%%%%%%%%%%%%%%%%%%%%%%%%%%%%%%%%%%
% APPENDIX
%%%%%%%%%%%%%%%%%%%%%%%%%%%%%%%%%%%%%%%%%%%%%%%%%%%%%%%%%%%%%%%%%%%%%%%%%%%%%%%
%%%%%%%%%%%%%%%%%%%%%%%%%%%%%%%%%%%%%%%%%%%%%%%%%%%%%%%%%%%%%%%%%%%%%%%%%%%%%%%
\newpage
\appendix
\onecolumn

\section{Prompts in Experienments}
\label{app:prompt}
We provide prompts for collecting teacher's responses in Figure~\ref{fig:gen_gsm8k} (GSM8K) and Figure~\ref{fig:gen_mmlu} (MMLU-PRO).

\begin{figure*}[!ht]
    \begin{tikzpicture}
    \node [draw, rounded corners,
         text width=\linewidth-24pt,    % <---
         align=flush center, 
         inner sep=12 pt,
         fill=lightgray,
    ]%
    {
    \begin{minipage}{1\linewidth}
    \footnotesize{
Q: \\
There are 15 trees in the grove. Grove workers will plant trees in the grove today. After they are done, there will be 21 trees. How many trees did the grove workers plant today? \\
A: \\
Let's break this down step by step!\\
Step 1: There are 15 trees originally. \\
Step 2: Then there were 21 trees after some more were planted. \\
Step 3: So there must have been 21 - 15 = 6.\\
The answer is 6.\\

Q: \\
If there are 3 cars in the parking lot and 2 more cars arrive, how many cars are in the parking lot? \\
A: \\
Let's break this down step by step! \\
Step 1: There are originally 3 cars. \\
Step 2: 2 more cars arrive, 3 + 2 = 5. \\
The answer is 5.\\

Q: \\
Leah had 32 chocolates and her sister had 42. If they ate 35, how many pieces do they have left in total?\\
A: \\
Let's break this down step by step! \\
Step 1: Originally, Leah had 32 chocolates. \\
Step 2: Her sister had 42. \\
Step 3: So in total they had 32 + 42 = 74. \\
Step 4: After eating 35, they had 74 - 35 = 39. \\
The answer is 39.\\

Q: \color{orange}{\{question}\} } \\
Let's break this down step by step!
\end{minipage}
};
\end{tikzpicture}
\caption{
Prompt template for generating responses in the teacher LLMs over GSM8K dataset.
}
\label{fig:gen_gsm8k}
\end{figure*}

\begin{figure*}[!ht]
    \begin{tikzpicture}
    \node [draw, rounded corners,
         text width=\linewidth-24pt,    % <---
         align=flush center, 
         inner sep=12 pt,
         fill=lightgray,
    ]%
    {
    \begin{minipage}{1\linewidth}
    \footnotesize{
Question:  {\color{orange}{\{question of fewshot example 1\}}} \\
Options: {\color{orange}{\{options of fewshot example 1\} }} \\
Answer: {\color{orange}{\{answer of fewshot example 1\} }} \\

Question:  {\color{orange}{\{question of fewshot example 2\} }} \\
Options: {\color{orange}{\{options of fewshot example 2\} }} \\
Answer: { \color{orange}{\{answer of fewshot example 2\} }}\\

Question: {\color{orange}{\{question\} }} \\
Options: { \color{orange}{\{options\}}}  \\
Answer: {} \\
Let's break this down step by step! }\\
\end{minipage}
};
\end{tikzpicture}
\caption{
Prompt template for generating responses in the teacher LLMs over MMLU-PRO dataset. We use two shots (provided by the dataset) for few-shot learning.
}
\label{fig:gen_mmlu}
\end{figure*}

\section{Implementation Details}
\subsection{Data Split}
\label{app:data_split}
For GSM8K, we divided the original training dataset into training and validation sets, allocating 90\% for training and 10\% for validation.

For MMLU-PRO, we first allocate 15\% of the data for testing. Then, we split the remaining data into training and validation sets using a 90\% to 10\% ratio.

\subsection{Hyperparameter}
\label{app:hyperparameters}
Our training pipeline consists of three stages: supervised fine-tuning (SFT) warm-up, reward model training, and proximal policy optimization (PPO). Each stage plays a critical role in progressively improving the student model.

\paragraph{Dataset Specific Hyper-parameters} We set the maximum generation length as $512$ for GSM8K and $1024$ for MMLU-PRO.
\paragraph{SFT Warm-up}

% Optimized Paragraph
The \textbf{Supervised Fine-Tuning (SFT)} phase serves to initialize the student model prior to reinforcement learning. During SFT, we employ a learning rate of \(5 \times 10^{-6}\) and a sequence length of 512 tokens. The batch size varies based on the specific teacher and student model configurations, as detailed in \textbf{Table \ref{tab:batch_sizes}}. Our dataset comprises majority-voted responses, ensuring a robust foundation for subsequent optimization. For the warm-up phase, we utilize 4 H100 GPUs and perform full parameter training. The training process spans 4 epochs, with checkpoints saved at intervals specified in the table. The optimal checkpoint is selected based on performance on the validation set. To accelerate training, we leverage DeepSpeed.

% Revised Table
\begin{table}[ht]
    \centering
    \begin{tabular}{ccccc}
        \toprule
        Dataset & Teacher Model & Student Model & Batch Size & Save Steps \\
        \midrule
        \multirow{4}{*}{GSM8K} 
            & \multirow{2}{*}{Llama3-70B} & Llama3-1B & 84 & 100 \\
            &                             & Llama3-3B & 74 & 100 \\ \cline{2-5}
            & \multirow{2}{*}{Llama3-8B} & Llama3-1B & 84 & 100 \\
            &                             & Llama3-3B & 70 & 100 \\
        \midrule
        \multirow{4}{*}{MMLU-PRO} 
            & \multirow{2}{*}{Llama3-70B} & Llama3-1B & 40 & 400 \\
            &                             & Llama3-3B & 32 & 400 \\ \cline{2-5}
            & \multirow{2}{*}{Llama3-8B} & Llama3-1B & 40 & 100 \\
            &                             & Llama3-3B & 32 & 100 \\
        \bottomrule
    \end{tabular}
    \caption{Batch size and checkpoint saving steps in warm up phase.}
    \label{tab:batch_sizes}
\end{table}

\paragraph{Reward Model Training}
The reward model is trained to guide PPO-based fine-tuning. This stage uses a learning rate of $5\times10^{-5}$, a batch size of 48 for student Llama3-1B and a batch size of 16 for student Llama3-3B, and 4 training epochs. We apply early stop while the reward model performance stop increasing on validation set.The reward model is initialized from the student model after warm up.
All reward models were trained on four H100 GPUs.

\paragraph{PPO Training}
The PPO stage refines the student model through reinforcement learning with the reward model. We use a learning rate of $1\times10^{-5}$, a KL penalty coefficient of 0.2, and a value function coefficient of 0.1. The total number of training episodes is set to $200,000$, ensuring sufficient interaction with the reward model for stable policy improvement. We apply early stop while the student model performance stop increasing on validation set. We present more hyper-parameters in Table~\ref{tab:batch_sizes_ppo}.

\begin{table}[ht]
    \centering
    \begin{tabular}{cccccc}
        \toprule
        Dataset & Teacher Model & Student Model & Batch Size & Learning Rate & GPU\_NUM\\
        \midrule
        \multirow{4}{*}{GSM8K} 
            & \multirow{2}{*}{Llama3-70B} & Llama3-1B & 20 & $1\times 10^{-5}$ & 2\\
            &                             & Llama3-3B & 4 & $5\times 10^{-6}$  & 4\\
            & \multirow{2}{*}{Llama3-8B} & Llama3-1B & 20 & $1\times 10^{-5}$  & 2\\
            &                             & Llama3-3B & 4 & $5\times 10^{-6}$  & 4\\
        \midrule
        \multirow{4}{*}{MMLU-PRO} 
            & \multirow{2}{*}{Llama3-70B} & Llama3-1B & 10 & $5\times10^{-6}$ &  4 \\
            &                             & Llama3-3B & 2 & $1\times10^{-5}$ & 4 \\
            & \multirow{2}{*}{Llama3-8B} & Llama3-1B & 10 & $1\times10^{-5}$  & 4 \\
            &                             & Llama3-3B & 2 & $1\times10^{-5}$  & 4  \\
        \bottomrule
    \end{tabular}
    \caption{Hyper-parameters in PPO.}
    \label{tab:batch_sizes_ppo}
\end{table}

\paragraph{\textbf{Ours w/o Data}} We apply a learning rate of $5\times^10{-5}$ and KL coefficient of 0.1 in this ablation version. Other hyper-parameters are the same to \textbf{PPO Training}.

%%%%%%%%%%%%%%%%%%%%%%%%%%%%%%%%%%%%%%%%%%%%%%%%%%%%%%%%%%%%%%%%%%%%%%%%%%%%%%%
%%%%%%%%%%%%%%%%%%%%%%%%%%%%%%%%%%%%%%%%%%%%%%%%%%%%%%%%%%%%%%%%%%%%%%%%%%%%%%%

\end{document}